\documentclass[10pt,twocolumn,letterpaper，threeparttable]{article}

\usepackage{wacv}
\usepackage{times}
\usepackage{color}
\usepackage{epsfig}
\usepackage{graphicx}
\usepackage{amsmath}
\usepackage{amssymb}
\usepackage{subfigure}
\usepackage{adjustbox}
\usepackage{algorithm}
\usepackage{algpseudocode}
\usepackage{booktabs} 

\usepackage[pagebackref=true,breaklinks=true,colorlinks,bookmarks=false]{hyperref}

\wacvfinalcopy 

\def\wacvPaperID{961} 

\ifwacvfinal\fi
\begin{document}

\title{Detecting Anti-Semitic Hate Speech using Transformer-based Large Language Models }

\author{Dengyi Liu, Minghao Wang, Andrew G Catlin\\
Katz School of Science and Health\\
Yeshiva University\\
{\tt\small dliu6@mail.yu.edu,     mwang3@mail.yu.edu,    andrew.catlin@yu.edu}
}

\maketitle
\ifwacvfinal\thispagestyle{empty}\fi

\begin{abstract}

Academic researchers and social media entities grappling with the identification of hate speech face significant challenges, primarily due to the vast scale of data and the dynamic nature of hate speech. Given the ethical and practical limitations of large predictive models like ChatGPT in directly addressing such sensitive issues, our research has explored alternative advanced transformer-based and generative AI technologies since 2019. Specifically, we developed a new data labeling technique and established a proof of concept targeting anti-Semitic hate speech, utilizing a variety of transformer models such as BERT~\cite{bert}, DistillBERT~\cite{distillbert}, RoBERTa~\cite{roberta}, and LLaMA-2~\cite{llama2}, complemented by the LoRA~\cite{lora} fine-tuning approach. This paper delineates and evaluates the comparative efficacy of these cutting-edge methods in tackling the intricacies of hate speech detection, highlighting the need for responsible and carefully managed AI applications within sensitive contexts.

\end{abstract}

\section{Introduction}



In our research, we investigate the prevalent and pernicious issue of anti-Semitic hate speech proliferating across digital platforms, particularly Twitter. This study is twofold: firstly, it involves the collection of online discourse related to anti-Semitic sentiment, upon which we apply our specially designed data labeling algorithm for annotation; secondly, we employ a variety of modeling techniques to discern and categorize hate speech. These methodologies include different embedding approaches coupled with machine learning algorithms such as k-NN (k-nearest neighbors), linear regression, random forest, and naive Bayes, in addition to state-of-the-art transformer-based large pre-trained models like BERT~\cite{bert}, DistilBERT~\cite{distillbert}, and RoBERTa~\cite{roberta}. Moreover, we refine the capabilities of RoBERTa~\cite{roberta} and LLaMA-2~\cite{llama2} through fine-tuning via Low-Rank Adaptation (LoRA)~\cite{lora} technology, aiming to significantly enhance their performance in detecting subtle and overt forms of anti-Semitic content.

The surge of anti-Semitic expression on social media platforms, unfortunately, parallels the rise in their user base, presenting multifaceted challenges to effective detection and moderation. By leveraging models recognized for their profound linguistic understanding, our work aims to amplify the efficacy of detection mechanisms against anti-Semitic expressions online. Through the application of sophisticated Natural Language Processing (NLP) techniques, this research endeavors to bolster online safety and confront the spectrum of hate speech. It provides valuable insights into how advanced NLP techniques can be intricately applied to elevate content moderation standards on digital platforms, thereby contributing to the critical conversation on combating online hate speech and ensuring a safer communicative environment.

\section{Related Work}\label{sec:related}
The battle against online hate speech has leveraged various NLP technologies to sift through vast online content, seeking patterns and markers indicative of harmful speech. This section reviews the literature on hate speech detection, with a focus on the evolution from rule-based and traditional machine learning approaches to the more recent adoption of transformer-based models.

\subsection{Embeddings in NLP}
The utilization of embeddings in NLP represents a critical evolution in how machines understand and process human language. Embeddings transform discrete textual elements—such as words, phrases, or sentences—into continuous vectors of real numbers, capturing semantic meanings in a dense representation. This technique significantly differed from earlier vectorization methods like one-hot encoding, which failed to encapsulate contextual relationships. Pioneering models such as Word2Vec and GloVe introduced by Mikolov et al. (2013) and Pennington et al. (2014) respectively, demonstrated how embeddings could capture syntactic and semantic word relationships based on their co-occurrence in large text corpora, fundamentally enhancing the machine's ability to interpret text~\cite{mikolov2013word2vec}\cite{pennington2014glove}. The development of contextual embeddings by models like ELMo and GPT further refined this approach by considering the context of words within sentences, leading to improvements in numerous NLP tasks\cite{peters2018elmo}~\cite{radford2018gpt}.

Traditional vectorization techniques play a vital role in feature extraction from text. Techniques such as CountVectorizer, TfidfVectorizer, and HashingVectorizer are fundamental in transforming raw text into numerical vectors that can be ingested by machine learning models. CountVectorizer converts text documents into a matrix of token counts, effectively capturing the frequency of each word within the documents. On the other hand, TfidfVectorizer combines the benefits of CountVectorizer with the TF-IDF (Term Frequency-Inverse Document Frequency) weighting, which diminishes the impact of tokens that occur very frequently and are thus less informative. HashingVectorizer, meanwhile, offers a stateless operation where the text is converted into a numeric hash value; this is particularly useful for datasets too large to fit in memory, as it does not require vocabulary fitting~\cite{scikitlearn}.

\begin{itemize}
\item \textbf{CountVectorizer}: Transforms text into a frequency matrix of tokens which can effectively support models in learning from text data.
\item \textbf{TfidfVectorizer}: Calculates TF-IDF scores for the tokens in text, balancing the frequency of tokens against their rarity across all documents, enhancing the ability to identify meaningful words.
\item \textbf{HashingVectorizer}: Applies a hashing function to term frequency counts, making it efficient for large-scale text processing as it avoids the need to fit and store a vocabulary in memory.
\end{itemize}

These vectorization techniques are instrumental in preprocessing text for both traditional machine learning and deep learning models, serving as a bridge between raw data and the sophisticated algorithms designed to process them.

\subsection{Traditional Methods to Hate Speech Detection}
Initial efforts in hate speech detection predominantly utilized rule-based algorithms and traditional machine learning models. These approaches, while foundational, often struggled with the subtleties of language and context, leading to high false positive rates and limited scalability. Seminal works such as Warner and Hirschberg (2012) and Davidson et al. (2017) explored the use of Support Vector Machines (SVM) and Naive Bayes classifiers, setting a precedent for automated content moderation but highlighting the need for more nuanced understanding and classification of text~\cite{warner2012svm}~\cite{davison2017svm}.

Further extending the repertoire of traditional machine learning methods, Logistic Regression, k-Nearest Neighbors (k-NN), and Random Forests have also been applied to hate speech detection with varying degrees of success. Logistic Regression, a linear model, is often favored for its simplicity and interpretability, making it particularly useful in scenarios where understanding the influence of individual features (such as specific words or phrases) on the prediction is crucial~\cite{haddad2019benchmarking}. Meanwhile, k-NN has been employed for its non-parametric nature, where the classification of a document is based on the majority label among its nearest neighbors in the feature space. This method is intuitive and effective, especially when dealing with well-segmented data clusters~\cite{zhang2018detecting}.

Random Forests, as an ensemble method comprising numerous decision trees, provide a robust alternative to single estimators like SVM. The strength of Random Forests lies in their ability to reduce overfitting by averaging multiple deep decision trees, trained on different parts of the same training set. This method is exceptionally good at handling the non-linear relationships and interactions within textual data, thus providing better generalization in detecting hate speech across diverse contexts~\cite{burnap2016us}.

These traditional models, though sometimes overshadowed by the more recent deep learning approaches, continue to be relevant, particularly in resource-constrained environments or as part of hybrid systems that incorporate both traditional and modern techniques.

\subsection{Deep Learning in NLP}
The introduction of deep learning to NLP marked a significant turning point, offering models capable of understanding complex language patterns through layered representations~\cite{hochreiter1997long}. Convolutional Neural Networks (CNNs) and Recurrent Neural Networks (RNNs) began to outperform their predecessors in various tasks, including sentiment analysis and topic classification. However, these models still faced challenges in capturing long-term dependencies and context in text, a crucial aspect of detecting hate speech~\cite{kim2014convolutional}~\cite{cho2014learning}.

\subsection{Transformer Architecture}
Since the invention of the transformer architecture~\cite{transformer2017}, the attention mechanism has been increasingly used in the NLP field due to its good parallel computing capabilities. Also, a transformer-based model can be used to do text classification tasks. The advent of transformer architecture revolutionized NLP by enabling models to process words concerning all other words in a sentence, thus capturing the context more effectively than ever before~\cite{transformer2017}. Building upon this, BERT (Bidirectional Encoder Representations from Transformers), developed by Devlin et al. (2019), represented a landmark in pre-trained models, allowing for an even more nuanced understanding of language through bidirectional training~\cite{bert}.

Following the success of BERT, DistilBERT was introduced by Sanh et al. (2019) as a smaller, faster, and lighter version of BERT. DistilBERT optimizes BERT’s architecture to maintain 97\% of its language understanding capabilities while reducing the model size by 40\% and speeding up training significantly~\cite{distillbert}. This makes it more accessible and practical for real-time applications.

RoBERTa (Robustly Optimized BERT Approach), developed by Liu et al. (2019), extends BERT by modifying key hyperparameters, training with much larger mini-batches and learning rates, and training on more data. RoBERTa removes BERT's next-sentence pretraining objective and dynamically changes the masking pattern applied to the training data. This has led to improved performance across a range of NLP tasks compared to BERT~\cite{roberta}.

An interesting advancement in model adaptation is LoRA (Low-Rank Adaptation), which can be applied to models like RoBERTa. LoRA allows for efficiently adapting large pre-trained models with minimal additional parameters, preserving the original parameters while achieving comparable performance to full fine-tuning approaches~\cite{lora}.

Recently, LLaMA-2 was introduced as an evolution of transformer models designed for generative and multitasking capabilities across multiple languages. LLaMA-2 benefits from a large-scale training corpus and has shown significant advancements in understanding and generating language tasks~\cite{llama2}. When combined with LoRA, LLaMA-2 can be fine-tuned in a parameter-efficient manner, allowing for custom adaptations without extensive retraining of the model.

These developments reflect the ongoing innovation within transformer-based architectures, making them more efficient and adaptable for a wide range of NLP applications.

\subsection{Application to Hate Speech Detection}
Following these advancements, several studies have specifically applied transformer-based models to the detection of hate speech. For instance, works by Mozafari et al. (2019) and Yuan et al. (2019) demonstrated the superior performance of BERT and its variants in classifying various forms of online abuse and hate speech, including anti-Semitic content. These studies underscore the effectiveness of transformer-based models in grasping the complexities and evolving nature of online discourse~\cite{mozafari2019bert}~\cite{yuan2019transfer}.

\subsection{Limitations and Future Directions}
Despite their success, transformer-based models are not without limitations. The high computational cost, need for extensive training data, and challenges in interpretability remain significant hurdles. Ongoing research is directed towards making these models more accessible and understandable, with efforts like DistilBERT and TinyBERT offering more efficient, albeit slightly less powerful, alternatives~\cite{distillbert}~\cite{jiao2019tinybert}.

\section{Methods}\label{sec:method}

\subsection{Data preparation and labeling method}

In the development of this research, the formation of a focused data analysis group was instrumental. This group was tasked with establishing a comprehensive strategy for data annotation, aimed at evaluating online discourse for potential anti-Semitic content. To this end, the Twitter Application Programming Interface (API) was employed to systematically retrieve a dataset consisting of approximately 10,000 posts, believed to have relevance to Jewish topics based on preliminary keywords and contextual analysis.

The data annotation process,
involved a detailed examination of the retrieved posts. This phase was critical in identifying and labeling the manifestations of anti-Semitic sentiments within the dataset. Through the collective 
efforts of the data team, a subset of 3,000 posts was labeled.

Following the initial phase of data collection and preliminary annotation, our research protocol detailed a structured approach for the in-depth analysis and classification of the data. The annotation process was carried out by a dedicated team of two researchers, who independently reviewed each post within the dataset. This methodological approach ensured that personal biases were minimized and that a comprehensive review of each post was conducted from multiple perspectives.

Upon completion of the independent review process by both team members, posts for which there was unanimous agreement—where both annotators identified a post as exhibiting anti-Semitic sentiment—were immediately assigned a final label. This approach streamlined the initial stage of data categorization, allowing for the rapid identification and classification of clear-cut cases of anti-Semitic content.

For posts where the two initial annotators disagreed, a distinctive procedure was implemented. These contentious cases were escalated to a third, impartial reviewer. This reviewer served as a mediator and provided an additional level of scrutiny. Following this, a joint discussion session was organized involving all three team members. During this collaborative review, the disputed posts were re-evaluated, and consensus was sought. The outcome of these discussions led to the final determination of the labels for these disputed posts, ensuring a thorough and democratically reached conclusion to each case of uncertainty. This meticulous approach to data annotation underscores the commitment of our research to precision and reliability in the identification of online hate speech.


\begin{algorithm}
\caption{Data Annotation with Threshold-Based Labeling and Dispute Resolution}
\begin{algorithmic}[1] 
\State \textbf{Input:} Dataset of posts ($D$)
\State \textbf{Output:} Annotated posts with labels ($L$)
\State Initialize annotated set $L = \emptyset$
\State Set scoring threshold $\theta = 6$  
\For{each post $p \in D$}
    \State Annotator1 scores $p$ with $score1$ from 0 to 10
    \State Annotator2 scores $p$ with $score2$ from 0 to 10
    \State Determine label for $p$ by Annotator1: $label1 = (score1 \geq \theta)$
    \State Determine label for $p$ by Annotator2: $label2 = (score2 \geq \theta)$
    \If{$label1 == label2$}
        \State Assign final label $label_f = label1$ to $p$
        \State Add $(p, label_f)$ to $L$
    \Else
        \State Assign $p$ to a third, impartial reviewer
        \State Third reviewer scores $p$ with $score3$ from 0 to 10
        \State Determine final label by third reviewer: $label_f = (score3 \geq \theta)$
        \State Add $(p, label_f)$ to $L$
    \EndIf
\EndFor
\State \textbf{return} $L$
\end{algorithmic}
\end{algorithm}

\begin{figure*}[h]
\centering
\includegraphics[width=2 \columnwidth]{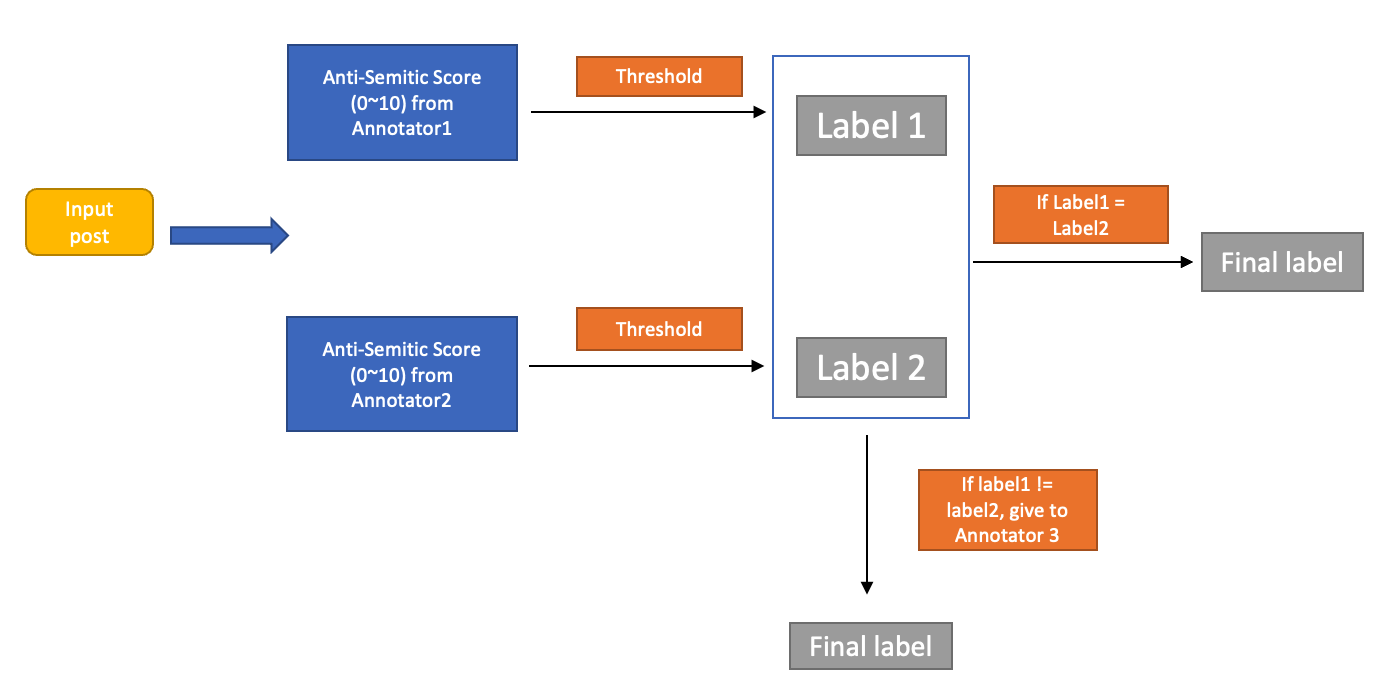}
\caption{This figure shows the workflow of voting algorithm.}
\label{fig:architecture}
\end{figure*}

\subsection{Train on machine learning and Fine-tuning transformer-based models}

In this segment of our research, we delineate the methodologies applied in training traditional machine learning models with different embeddings and fine-tuning transformer-based models to classify online discourse potentially exhibiting anti-Semitic content. This classification effort was premised on a well-defined data preparation and annotation process, as delineated in preceding sections.

Initially, a dataset comprising approximately 10,000 posts suspected to carry relevance to Jewish topics was systematically compiled using the Twitter API. A focused data analysis group undertook a meticulous annotation strategy, identifying and labeling instances of anti-Semitic sentiments within a subset of 3,000 posts. This foundational work laid the groundwork for the subsequent training and evaluation of various classification models.

Machine Learning Models Training: Traditional machine learning models including Naive Bayes, SVM, Random Forests, Logistic Regression, and K-Nearest Neighbors (K-NN) were employed. These models were trained using a standard approach, involving feature extraction techniques such as TF-IDF for text representation, followed by model-specific parameter tuning. The performance of these models, as outlined in Table \ref{final_result}, indicates varied levels of efficacy in classifying the annotated data, with SVM showing a balanced performance and K-NN lagging in all evaluated metrics.

Fine-tuning Transformer-based Models: In contrast to traditional machine learning approaches, transformer-based models, specifically BERT~\cite{bert}, DistilBERT~\cite{distillbert}, RoBERTa~\cite{roberta} and Llama-2 (7B)~\cite{llama2}, were fine-tuned for our classification task. These models leverage deep learning techniques that comprehend the complexities and nuances of natural language, offering superior performance over their machine learning counterparts. The fine-tuning process involved adjusting pre-trained models to our specific dataset, optimizing them to understand the linguistic subtleties of anti-Semitic discourse.

In traditional fine-tuning approaches for transformer-based models, extensive parameter adjustments are typically required, which can be computationally expensive and time-consuming, especially for models with a large number of parameters such as LLaMA-2 (7 billion parameters). To address this challenge, we implemented the Low-rank Adaptation (LoRA) method~\cite{lora} to fine-tune RoBERTa and LLaMA-2. LoRA strategically modifies only a small fraction of the model's parameters by introducing low-rank matrices that adapt the pre-trained weights rather than altering them extensively. This approach significantly reduces the computational overhead by limiting the scope of parameter updates during the training phase. Our experiments demonstrated that, by using LoRA, we were able to achieve comparable or even superior performance to traditional full-model fine-tuning while drastically cutting down on training time. This efficiency makes LoRA an exceptionally viable option for enhancing large-scale language models without the substantial resource investment typically associated with conventional fine-tuning techniques.

The results, as demonstrated, underscore the superiority of transformer-based models in handling the intricacies of language and sentiment in online hate speech. BERT, in particular, achieved remarkable results, affirming the effectiveness of fine-tuning approaches in enhancing model performance.

The distinctions in performance between machine learning models and transformer-based models highlight the advancements in NLP technology and their applicability in addressing complex societal issues like online hate speech. This analysis not only informs our understanding of the models' capabilities but also guides future efforts in online discourse classification.

In conclusion, the training on machine learning and fine-tuning of transformer-based models were instrumental in advancing our research objectives. The 
preparation and annotation of our dataset provided a robust foundation for evaluating these models, ultimately contributing to our overarching goal of identifying and mitigating anti-Semitic content online.

\begin{figure*}[h]
\centering
\includegraphics[width=1.5 \columnwidth]{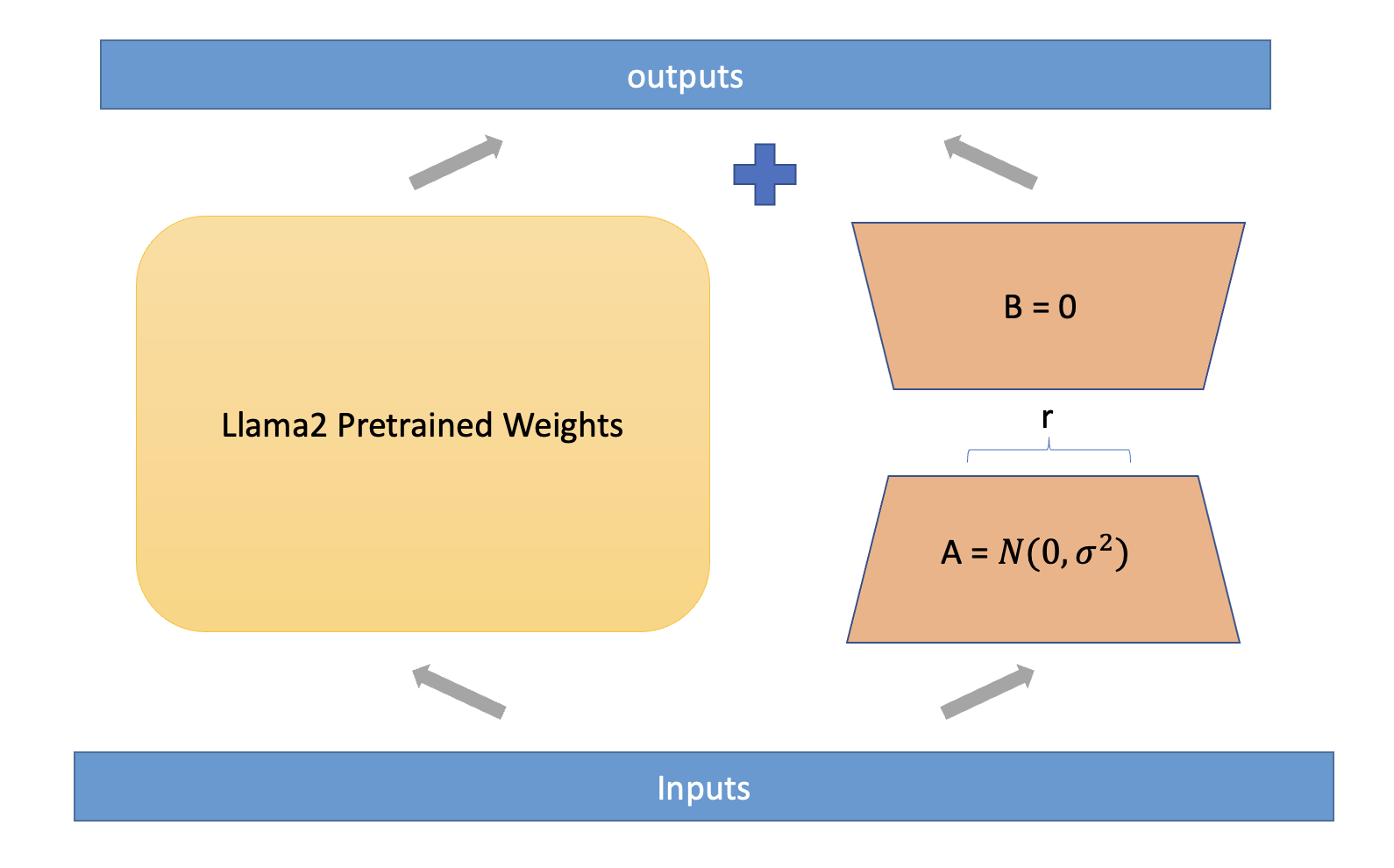}
\caption{ Architecture of LoRA fine-tuning applied to Llama2 model weights, demonstrating how the injection of trainable low-rank matrices A and B (initialized as A with a normal distribution and B set to zero) allows for efficient adaptation of the model's weights to new tasks, preserving the original parameters while introducing minimal updates.}
\label{fig:architecture}
\end{figure*}

\section{Results}\label{sec:results}

\begin{table*}[ht]
\centering
\caption{Results of Different Models}
\begin{tabular}{lccccc}
\hline
\textbf{Model} & \textbf{Accuracy} & \textbf{Precision} & \textbf{Recall} & \textbf{F1-score} & \textbf{Embedding} \\
\hline
Logistic Regression & 0.91 & 0.87 & 0.66 & 0.75 & CountVectorizer \\
 & 0.91 & 0.94 & 0.55 & 0.70 & TfidfVectorizer \\
 & 0.91 & 0.90 & 0.61 & 0.72 & HashingVectorizer \\
 & 0.81 & 0.50 & 0.18 & 0.26 & Word2Vec \\
 & 0.85 & 0.76 & 0.32 & 0.46 & GloVe \\
 \hline
k-NN & 0.85 & 0.79 & 0.30 & 0.44 & CountVectorizer \\
 & 0.91 & 0.91 & 0.58 & 0.71 & TfidfVectorizer \\
 & 0.91 & 0.93 & 0.56 & 0.70 & HashingVectorizer \\
 & 0.83 & 0.57 & 0.40 & 0.47 & Word2Vec \\
 & 0.86 & 0.87 & 0.35 & 0.50 & GloVe \\
 \hline
SVM & 0.90 & 0.79 & 0.69 & 0.73 & CountVectorizer \\
 & 0.92 & 0.91 & 0.67 & 0.77 & TfidfVectorizer \\
 & 0.92 & 0.90 & 0.67 & 0.77 & HashingVectorizer \\
 & 0.81 & 0.00 & 0.00 & 0.00 & Word2Vec \\
 & 0.85 & 0.80 & 0.31 & 0.44 & GloVe \\
 \hline
Random Forests & 0.91 & 0.96 & 0.55 & 0.70 & CountVectorizer \\
 & 0.91 & 0.96 & 0.55 & 0.70 & TfidfVectorizer \\
 & 0.91 & 0.96 & 0.55 & 0.70 & HashingVectorizer \\
 & 0.86 & 0.72 & 0.47 & 0.57 & Word2Vec \\
 & 0.86 & 0.87 & 0.33 & 0.48 & GloVe \\
 \hline
Naive Bayes & 0.91 & 0.94 & 0.56 & 0.70 & CountVectorizer \\
 & 0.89 & 1.00 & 0.46 & 0.63 & TfidfVectorizer \\
\hline
BERT & 0.94 & 0.96 & 0.73 & 0.81 & BertTokenizer \\
\hline
DistillBert & 0.92 & 0.91 & 0.71 & 0.80 & DistillBertTokenizer \\
\hline
Roberta & 0.93 & 0.88 & 0.75 & 0.81 & RoBertaTokenizer \\
RoBerta + lora & 0.91 & 0.86 & 0.73 & 0.79 & RobertaTokenizer \\
\hline
Llama-2 (7B)  & 0.93 & 0.79 & 0.78 & 0.82 & LlamaTokenizer \\
Llama-2 (7B) + lora & 0.92 & 0.80 & 0.79 & 0.83 & LlamaTokenizer \\
\hline
\end{tabular}
\end{table*}

\begin{table*}[ht]
\centering
\caption{Comparison of Fine-tuning time of different Transformer-based models}
\begin{tabular}{lcccc}
\hline
\textbf{Method} & \textbf{Fine-tuning Epochs} & \textbf{Platform(GPU)} & \textbf{Fine-tuning time}  \\ \hline
BERT & 30 epochs & Nvidia A4000(16GB) & 176 mins  \\
DistillBert  & 20 epochs & Nvidia A4000(16GB) & 79 mins  \\
RoBerta & 10 epochs & Nvidia A4000(16GB) & 54 mins  \\
RoBerta + lora & 10 epochs & Nvidia A4000(16GB) & 21 mins  \\
Llama-2 (7B) & 3 epochs & Nvidia A100(80GB) & 112 mins \\
Llama-2 (7B) + lora & 3 epochs & Nvidia A100(80GB) & 35 mins  \\ \hline
\end{tabular}

\label{final_result}
\end{table*}

Table 1 presents a comprehensive evaluation of various machine learning algorithms' performance with different text embeddings. The algorithms considered include Logistic Regression, k-NN, SVM, Random Forests, and Naive Bayes, assessed across metrics such as Accuracy, Precision, Recall, and F1-score. These metrics are pivotal for understanding the models' effectiveness in classification tasks, with each offering insights into different aspects of performance.

Logistic Regression shows consistent accuracy across all embeddings but exhibits varying degrees of precision, recall, and F1-scores, pointing its sensitivity to the choice of embedding. The traditional CountVectorizer, TfidfVectorizer, and HashingVectorizer perform comparably, whereas Word2Vec and GloVe show reduced effectiveness, particularly in recall and F1-score.

k-NN and SVM display a similar result, with TfidfVectorizer and HashingVectorizer providing optimal performance. Significantly, SVM shows a complete lack of recall and F1-score when utilizing Word2Vec, indicating a significant mismatch between this embedding and the SVM algorithm for the data in question.

Random Forests demonstrates a high degree of consistency across CountVectorizer, TfidfVectorizer, and HashingVectorizer, matching the highest precision observed but with moderate recall and F1-scores. This suggests Random Forests' robustness to different sparse representations while highlighting potential limitations with dense embeddings like Word2Vec and GloVe.

Naive Bayes is noteworthy for achieving a precision of 1.00 with TfidfVectorizer, albeit at the cost of reduced recall and F1-score. This notes that while Naive Bayes can be highly confident in its predictions, it may also miss a significant number of relevant instances. In addition, Naive Bayes used only two embeddings because other methods produce negative values, whereas naive bayes requires all inputs to have positive values.

All these results illustrate the importance of selecting appropriate embeddings for different algorithms. While some models like Random Forests and Naive Bayes exhibit robustness to the choice of embedding, others, such as SVM, show pronounced variability. These findings underscore the need for careful embedding selection in machine learning applications to optimize performance across all evaluation metrics.

Our analysis provides a comprehensive evaluation of various machine learning and deep learning models for the classification of online anti-Semitic discourse. The performance of each model, as summarized in Table~\ref{final_result}, is assessed based on Accuracy, Precision, Recall, and F1-score. 

The results indicate that transformer-based models, particularly BERT, demonstrate superior performance over traditional machine learning classifiers. BERT achieves the highest Accuracy and F1-score, highlighting the effectiveness of deep learning models that capture contextual nuances in text. Meanwhile, fine-tuning with LoRA shows promise, especially in improving Recall for RoBERTa and F1-scores for LLaMA-2 (7B), suggesting that targeted optimizations can further enhance the performance of pre-trained models.

Traditional algorithms like Naive Bayes, SVM, Random Forests, and Logistic Regression, while effective to a degree, fall short of the transformer-based models in various metrics. Notably, K-NN exhibits the least effectiveness, indicating that simpler classification methods may struggle with the complexity of language inherent in hate speech detection.

In summary, our results advocate for the adoption of transformer-based models in the automated detection of anti-Semitic hate speech, with fine-tuning techniques such as LoRA offering additional improvements in model sensitivity and specificity.

\section{Discussion}\label{sec:dis}

This study has examined the
capabilities of both traditional machine learning and modern transformer-based models in identifying anti-Semitic content within online discourse. The annotation process, underpinned by a democratic approach, established a robust dataset for training and evaluating the performance of different models.

Our findings reveal that while traditional machine learning models, such as SVM and Logistic Regression, provide a decent baseline in terms of accuracy and precision, they fall short in capturing the nuanced context and complexities inherent in natural language. This is evident in the relatively low recall scores, indicating a significant number of relevant cases went undetected. Conversely, K-NN's underperformance across all metrics highlights the challenges traditional algorithms face with high-dimensional and sparse textual data.

The superior performance of transformer-based models, notably BERT, DistilBERT, and RoBERTa, underscores the transformative impact of deep learning in natural language processing tasks. These models achieved notably higher scores in all metrics, particularly in precision and recall, illustrating their enhanced ability to discern nuanced language patterns and contextual meanings. This is a critical advancement, considering the sophisticated and evolving nature of online hate speech.

However, the deployment of such models is not without challenges. Transformer-based models require significant computational resources and expertise, potentially limiting their accessibility for smaller organizations or independent researchers. Additionally, the black-box nature of these models can pose interpretability challenges, complicating the process of understanding and explaining the basis of their classifications.


\section{Conclusion}\label{sec:conclusion}
The research undertaken provides valuable insights into the application of machine learning and deep learning techniques for social good, specifically in combating anti-Semitic hate speech online. Our comprehensive approach, from data collection and annotation to model training and evaluation, highlights the nuanced differences between various classification methods.

The study confirms the advanced capability of transformer-based models in understanding and classifying complex language patterns, outperforming traditional machine learning models. This finding aligns with the broader trend in natural language processing and offers a promising avenue for future research and practical applications in online hate speech detection.

Nonetheless, the ethical implications, computational costs, and interpretability issues associated with these advanced models warrant careful consideration. Future work should focus on addressing these challenges, improving model accessibility, and enhancing transparency without sacrificing performance.

Ultimately, this research contributes to the ongoing effort to create safer online environments. By leveraging cutting-edge technology and rigorous methodologies, we move closer to understanding and mitigating the spread of hate speech on digital platforms.



{\small
\bibliographystyle{unsrt}
\bibliography{egbib}
}

\end{document}